\pdfoutput=1

\documentclass[11pt]{article}

\usepackage[final]{acl}

\usepackage{times}
\usepackage{latexsym}
\usepackage{svg}
\usepackage{adjustbox}
\usepackage[T1]{fontenc}
\usepackage[T1,T2A]{fontenc}

\usepackage[utf8]{inputenc}

\usepackage{microtype}

\usepackage{lipsum} 
\usepackage{arabtex}
\usepackage{algorithm}
\usepackage{algpseudocode}

\usepackage[english,vietnamese,russian,arabic]{babel}
\usepackage{utf8}
\setcode{utf8}
\usepackage{inconsolata}
\usepackage{microtype}
\usepackage{graphicx}
\usepackage{import}
\usepackage{layout}
\usepackage{tabularx, makecell}
\usepackage{booktabs}
\usepackage{mathrsfs}
\usepackage{amssymb} 
\usepackage{url}
\usepackage{hyperref}
\usepackage{graphicx}
\usepackage{xspace,paralist}
\usepackage{times,latexsym}
\usepackage{amsmath}
\usepackage{appendix}
\usepackage{comment} 
\usepackage{enumitem}
\usepackage{makecell}
\usepackage{multirow}
\usepackage{xcolor}
\usepackage{arydshln}
\usepackage{cleveref}
\usepackage{todonotes}
\usepackage{longtable,supertabular}

\usepackage{svg}
\usepackage{amssymb}
\usepackage{pifont}
\usepackage{CJKutf8}

\usepackage[utf8]{inputenc}

\title{MuDRiC: Multi-Dialect Reasoning for Arabic Commonsense Validation}

\author{
  Kareem Elozeiri$^{*}$, Mervat Abassy$^{*}$, Preslav Nakov, Yuxia Wang \\
  Mohamed bin Zayed University of Artificial Intelligence, Abu Dhabi, UAE \\
  \texttt{\{kareem.ali , mervat.abassy\}@mbzuai.ac.ae}  
}

\begin{document}

\selectlanguage{english}
\maketitle
\def\thefootnote{*}\footnotetext{Equal contribution.}\def\thefootnote{\arabic{footnote}}

\begin{abstract}
Commonsense validation evaluates whether a sentence aligns with everyday human understanding, a critical capability for developing robust natural language understanding systems. While substantial progress has been made in English, the task remains underexplored in Arabic, particularly given its rich linguistic diversity. Existing Arabic resources have primarily focused on Modern Standard Arabic (MSA), leaving regional dialects underrepresented despite their prevalence in spoken contexts. To bridge this gap, we present two key contributions. We introduce MuDRiC, an extended Arabic commonsense dataset incorporating multiple dialects. To the best of our knowledge, this is the first Arabic multi-dialect commonsense reasoning dataset.
We further propose a novel method adapting Graph Convolutional Networks (GCNs) to Arabic commonsense reasoning, which enhances semantic relationship modeling for improved commonsense validation. Our experimental results demonstrate that this approach consistently outperforms the baseline of direct language model fine-tuning. Overall, our work enhances Arabic natural language understanding by providing a foundational dataset and a new method for handling its complex variations. Data and code are available at \textit{https://github.com/KareemElozeiri/MuDRiC}.



\end{list} 
\end{abstract}


\section{Introduction}
Common sense reasoning is a fundamental task in natural language processing (NLP), enabling machines to interpret and generate text in ways that align with human intuition~\citep{sap-etal-2020-commonsense}. It is critical for AI systems to make plausible inferences about the world and engage in human-like conversation~\citep{10.1145/2701413}. However, conversational commonsense often involves implicit social norms, cultural references~\citep{sadallah-etal-2025-commonsense}, and pragmatic reasoning that vary across dialects. Despite progress in English~\citep{10.5555/3031843.3031909, 10.1609/aaai.v33i01.33013027, talmor-etal-2019-commonsenseqa} and other high-resource languages, common sense reasoning remains challenging for languages with dialectal diversity, such as Arabic, primarily due to severe scarcity of annotated data for dialects. 

Most existing Arabic common sense benchmarks focus exclusively on Modern Standard Arabic (MSA), neglecting the rich diversity of Arabic dialects~\citep{lamsiyah-etal-2025-arabicsense, sadallah-etal-2025-commonsense, 10453697}. Dialects such as Egyptian, Gulf, Levantine, and Moroccan dominate everyday communication across the Arab world. Beyond lexical or grammatical variation, these dialects encode fine-grained regional cultural knowledge, making dialectal commonsense reasoning a culturally grounded and challenging task. As a result, models trained solely on MSA often fail to generalize to dialectal content. To address this gap, we introduce the first multi-dialect Arabic commonsense dataset balanced across Egyptian, Gulf, Levantine, and Moroccan dialects.

In terms of approaches to address Arabic common sense tasks, prior work heavily relies on MSA-centered models, e.g.,\ AraBERT~\citep{antoun-etal-2020-arabert}, which perform barely above chance on dialectal data~\citep{lamsiyah-etal-2025-arabicsense, 10453697}. Dialect-specific models such as MARBERT~\citep{abdul-mageed-etal-2021-arbert} also show weak performance due to nuanced differences between dialects. More related work in Appendix~\ref{sec:relatedwork}.

We propose integrating base language models with graph-based augmentation to capture deeper semantic relationships. This integration of graph-based methods significantly enhances cross-dialect robustness. To summarize our main contributions:
\begin{itemize}
    \item We introduce MuDRiC: the first multi-dialect Arabic common sense benchmark, enabling more inclusive and robust Arabic NLP systems. 
    \item We propose graph-based augmentation training strategy to enhance performance on dialectical data.
\end{itemize}
\section{Dataset}


\paragraph{Task Description and Formulation} 
Given a single sentence, the task aims to identify whether it is reasonable (labeled as 1) or non-reasonable (labeled as 0), based on its alignment with common sense. We cast commonsense validation as a binary classification task to provide a unified and simple formulation across datasets with different original structures. This setting allows us to evaluate the commonsense plausibility of a single sentence independently, rather than relying on relative comparisons between candidates and facilitates scaling the task to multiple Arabic dialects. 
\paragraph{MSA} We use two established Modern Standard Arabic (MSA) datasets for commonsense validation: the Arabic Dataset for Commonsense Validation (ADCV; \citet{tawalbeh2020sentence}) and ArabicSense~\citep{lamsiyah-etal-2025-arabicsense}.ADCV contains 11,000 instances, each consisting of a pair of sentences, where the task is to select the more commonsensical option. We convert this setup into a binary classification task by separating each sentence pair into two individual sentences, assigning label 1 to the original correct (reasonable) sentence and label 0 to its incorrect counterpart. This results in 22,000 labeled samples. ArabicSense includes 5,650 multiple-choice instances with two candidate sentences per instance, one of which is commonsensical. We apply the same conversion strategy, assigning labels accordingly, yielding 11,288 MSA samples after removing duplication.

\paragraph{Dialects Extension} 

Based on the MSA datasets above, we translate them into four Arabic dialects including Egyptian, Moroccan, Gulf and Levantine using GPT-4o~\citep{openai2024gpt4o}. The statistical distribution of the extended datasets is summarized in Table~\ref{tab:data-distribution}, while Table~\ref{tab:msa-dialect-examples} presents sample MSA sentences from ADCV alongside their corresponding dialectal translations. Prompting details can be found in Appendix \ref{datagen}.

The final composite dataset ensures a parallel representation across four major Arabic dialect families. This addresses a critical gap in Arabic NLP, where previous benchmarks have been limited to either Modern Standard Arabic or isolated dialectal efforts without systematic comparison.


\label{sec:data-annotation}
\begin{table}[t!]
\centering
\adjustbox{max width=\linewidth}{
\begin{tabular}{lcc}
\toprule
\textbf{Source Dataset} & \textbf{MSA Samples} & \textbf{Dialectal Samples} \\
\midrule
\textbf{ADCV} & 22,000 & 88,000 \\
\textbf{ArabicSense} & 11,288 & 45,152 \\
\midrule
\textbf{Total} & \textbf{33,288} & \textbf{133,152} \\
\bottomrule 
\end{tabular}
}
\caption{Statistical distribution of datasets, with each MSA extending to four dialects.}
\label{tab:data-distribution}
\end{table}

\begin{table*}[t!]
    \centering
    \adjustbox{max width=\textwidth}{
    \begin{tabular}{@{}p{6cm}p{6cm}p{6cm}p{6cm}p{6cm}p{1cm}@{}}
        \toprule
        \textbf{MSA Text } & \textbf{Egyptian} & \textbf{Gulf} & \textbf{Moroccan} & \textbf{Levantine} & \textbf{Label} \\ 
        \midrule
        \makebox[5cm][r]{\RL{لا أحد يريد العيش مع الفئران}} \newline \newline (No one wants to live with rats) & \makebox[5cm][r]{\RL{محدش عايز يعيش مع الفيران}} & \makebox[5cm][r]{\RL{ما في أحد يبي يعيش مع الفيران}} & \makebox[5cm][r]{\RL{حتى واحد ما بغا يعيش مع الفيران}} & \makebox[5cm][r]{\RL{ما حدا بده يعيش مع الفيران}} & 1 \\ 
        \midrule
        \makebox[5cm][r]{\RL{تقوم جورجيا تك بتدريب التنين}} \newline \newline (Georgia Tech trains dragons) &\makebox[5cm][r]{\RL{جورجيا تك بتدرب التنين}} &\makebox[5cm][r]{\RL{جورجيا تك تقوم بتدريب التنين}} & \makebox[5cm][r]{\RL{جورجيا تك كيدرّبو التنين}} &  \makebox[5cm][r]{\RL{جورجيا تك عم تدرب التنين}} & 0 \\ 
        \bottomrule
    \end{tabular}
    }
    \caption{MSA and dialectal examples from ADCV. Label 1 = reasonable and label 0 = non-reasonable sentence.}
    \label{tab:msa-dialect-examples}
\end{table*}
\paragraph{Quality Control}
We apply multiple quality control steps to ensure the reliability of the translated dataset. First, each dialectal translation is automatically verified using Gemini 2.5 Flash by jointly evaluating the translated sentence and its original MSA counterpart (Prompt in Appendix \ref{datagen}). Samples flagged as incorrect by Gemini are subsequently reviewed by native-speaker annotators. In total, approximately 8,580 samples (5.2\% of the dataset) were flagged; among these, 27\% were confirmed to be incorrect (corresponding to roughly 1.4\% of the full dataset) and were corrected by the annotators.

As an additional validation step to the original source datasets, we randomly sample 500 instances and ask two independent annotators to verify the correctness of their commonsense labels (reasonable vs. non-reasonable). All sampled instances were found to be correctly labeled.

\section{Methodology}
\label{sec:models}

We explore (i) graph-based augmentation to inject relational structure and (ii) domain-adversarial training (Appendix~\ref{append:adversarial}) to encourage dialect-invariant representations.


\subsection{Graph-based Language Model Reps}

Inspired by prior work that integrates graph encoders with Transformer models~\citep{Jiawei-etal-2020-graphbert,lu-etal-2020-vgcnbert}, we augment pretrained Masked Language Models (MLMs) with a Graph Convolutional Network (GCN) that encodes local word-level relations and surface morphological cues. This is motivated by the continuum of Arabic dialects and their non-standard orthography, which introduce substantial spelling and morphological variation. As a result, sequence-based fine-tuning becomes brittle, and dialect-invariant objectives are less reliable~\citep{shaban-habash-2025-arabic,bhatia-etal-2025-swan}. The graph encoder connects related variants via message passing, complementing contextual semantics and improving cross-dialect commonsense validation.

\begin{figure*}[t]
    \centering
        \includegraphics[width=\linewidth]{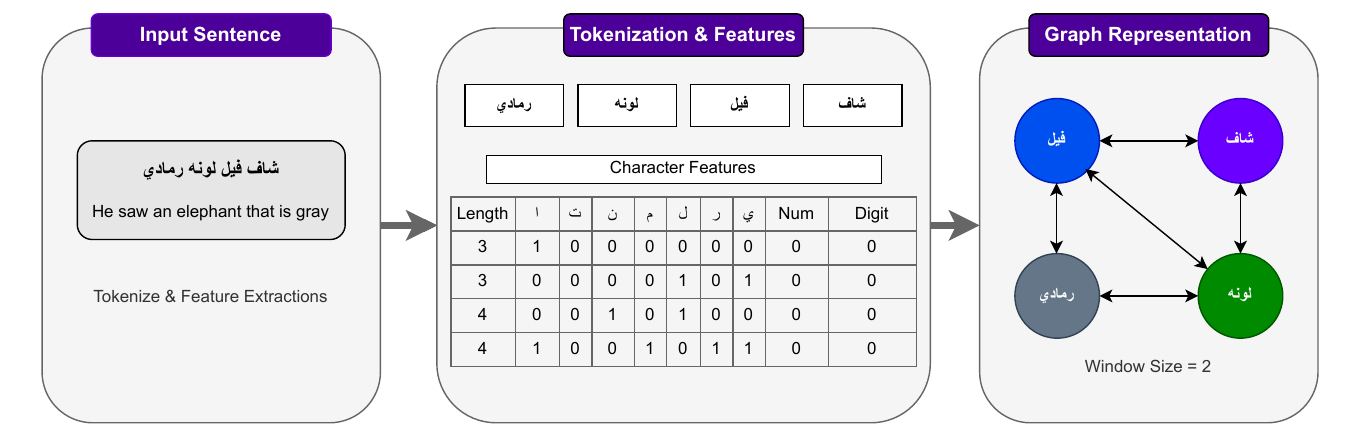}
    \caption{The creation process of the graph representation for a sentence.}
    \label{fig:graph_creation}
\end{figure*}


\paragraph{Pipeline Overview}
We first construct a semantic graph for each input instance to capture relational dependencies between entities. This graph is processed using GCNs to produce a fixed-length graph representation. In parallel, a BERT-based encoder generates a contextualized text embedding, which is projected to the same dimensionality as the graph representation. The two embeddings are then fused via concatenation. This fusion enriches the model by combining token-level semantic representations with higher-order structural information, enabling more effective reasoning over the input.

\paragraph{Graph Representation}

To build the graph representation (Figure~\ref{fig:graph_creation}), each input text is first tokenized into words. A co-occurrence graph is then constructed, where nodes correspond to unique words and undirected edges connect words appearing within a fixed-size sliding window. Each node is initialized with a handcrafted feature vector derived from word-level statistics, including word length and Arabic-specific morphological indicators (e.g., character counts and digit-related features). This results in lightweight and informative node representations that reflect the surface morphology and character patterns in the Arabic text.

The resulting graph is then processed by a multilayer GCN with one hidden layer. The GCN layers propagate and aggregate features across the graph, enabling the model to learn contextual structural patterns. A global mean-pooling layer is then applied to extract a single fixed-length vector that summarizes the entire graph.

\paragraph{Semantic Representation by MLMs}
In parallel, the input text is encoded using a BERT-based model. We extract the contextual embedding of the ``[CLS]'' token from the final hidden state layer, which serves as a summary representation of the input sequence. Both the graph and the BERT embeddings are projected into a shared fusion space using learnable linear projections.

\paragraph{Fusion} 
To combine the two representation inputs, we employ a multi-head self-attention mechanism over the concatenated graph and MLM embeddings. This allows the model to dynamically weigh the contribution of each modality and to learn complex interactions between them. The output of the attention layer is flattened and passed through a feedforward classification head. Figure \ref{fig:model_arch} and Algorithm \ref{algo:training-graphTransformer} summarize the pipeline in Appendix \ref{append:graph}.

\section{Experiments}
\label{sec:exp}

\begin{table*}[t!]
\centering
\small
\begin{tabular}{lcccccc}
\toprule
\textbf{Methods} & \textbf{MSA} & \textbf{Egyptian} & \textbf{Gulf} & \textbf{Levantine} & \textbf{Moroccan} & \textbf{Avg.} \\ 
\midrule
AraBERTv2  & 65.88 & 68.33& 65.78& 65.36& 62.09& 65.34 \\
AraBERTv2 + GCN &  \textbf{68.15} & \textbf{70.00}& \textbf{66.63}& \textbf{67.03 }& \textbf{63.24} & \textbf{67.01} \\
\midrule
CAMeLBERT-mix &  73.30 & 73.52 & 74.63& 73.82 & 66.45 & 72.34 \\
CAMeLBERT-mix + GCN &  \textbf{74.21}& \textbf{75.61} & \textbf{76.24} & \textbf{76.70}& \textbf{68.87} & \textbf{74.28}\\
\midrule
MARBERTv2  & 80.12& 80.73& 78.81& 80.03& 71.09 & 78.16 \\
MARBERTv2 + GCN & \textbf{81.64}& \textbf{81.26} & \textbf{80.87} & \textbf{80.64}& \textbf{73.06} & \textbf{79.53} \\

\bottomrule
\end{tabular}
\caption{Accuracy of baselines and our method on MSA and dialects datasets in \% (higher is better).}
\label{tab:acc-results}
\end{table*}

\subsection{Experimental Setup}
\label{sec:exp-setup}
\paragraph{Baselines}
We fine-tune three Arabic-centric pretrained masked language models (MLMs), i.e.,\ CAMeLBERT-mix \citep{inoue-etal-2021-interplay}, AraBERT and MARBERT as baselines, and evaluated on all dialects datasets. 
The three models were pretrained on large Arabic corpus, making them suitable for the task at hand. AraBERT focuses on Modern Standard Arabic. MARBERT emphasizes dialectal Arabic, incorporating substantial representation from various regional dialects, which can better capture the dialectal characteristics of the dataset. Similar to MARBERT, CAMeLBERT-mix was pretrained to dialectal Arabic in addition to modern standard Arabic and classical Arabic. 

\paragraph{Our Methods}
We evaluated the effectiveness of 
graph-based representations of MLMs which fused GCN-based embeddings with masked language models' contextual embeddings. 
This experimental setup allowed us to examine the hypothesis that graph-enhanced representations can improve downstream task performance.

\paragraph{Data Split and Training Setups}
In all experiments, we trained models using cross-entropy loss and optimized with AdamW~\citep{Loshchilov2017DecoupledWD}, using a learning rate of  2e-5, weight decay of 0.01, and a batch size of 128 for 3 epochs. We used 110K MuDRiC samples with ADCV as source dataset, balanced across both commonsense and dialect labels. The data were split into 70\%/15\%/15\% train, development, and test sets. This split was fixed and reused across all experiments to ensure fair and consistent comparisons.

\subsection{Results}

Table \ref{tab:acc-results} reports accuracy on MSA and four dialectal subsets. The \textit{Avg. Dialects} column corresponds to the overall average across all five subsets (MSA + the four dialects).

\paragraph{Which is the Best Base Model?}
Across the table, performance follows a stable ranking: \textit{MARBERTv2} $>$ \textit{CAMeLBERT-mix} $>$ \textit{AraBERTv2}. This ordering is consistent with the expected degree of \emph{dialect exposure} during pretraining: MARBERTv2 is explicitly dialect-heavy, CAMeLBERT-mix is more balanced, and AraBERTv2 is comparatively more MSA-oriented. The implication is that dialectal generalization is primarily constrained by representation quality learned during pretraining; downstream methods help, but they do not compensate fully for a strong pretraining mismatch.


\paragraph{GCN Consistently Improves Performance.}
The consistent improvements across all models suggest that the GCN provides information that the transformer alone does not reliably capture, such as relational structure, lexical/semantic neighborhood effects, or instance-to-instance dependencies. Importantly, the gains appear larger (in relative terms) for the weaker backbones (AraBERTv2 and CAMeLBERT-mix), which supports the interpretation that GCN fusion helps \emph{mitigate} dialect/domain mismatch by encouraging useful sharing across related samples or features. For MARBERTv2, the improvement is still present but more incremental, consistent with the idea that a dialect-rich backbone already captures much of the needed variation and the GCN mainly refines decision boundaries.

Overall, the table supports a clear conclusion: dialect-aware pretraining is the strongest driver of performance, and GCN fusion is a reliable enhancement.
\section{Conclusion}
This work presented two major contributions to Arabic NLP: (1) the creation of the first large-scale, multi-dialect common sense reasoning dataset, and (2) Enhanced Arabic Commonsense Reasoning methodology combining  graph-based embeddings with pre-trained BERT-based models to enhance performance. By systematically expanding MSA commonsense reasoning benchmarks into four major dialects we established a crucial resource for evaluating dialect robustness. Our experiments demonstrated that neither MSA-focused models (e.g.,\ AraBERT) nor dialect-pretrained models (e.g., MARBERT) alone suffice for reliable common sense classification across dialects. Instead, our hybrid graph-based approach to structured commonsense representation outperformed prior methods, setting a new benchmark for dialect-aware Arabic NLP.


\section*{Limitations \& Future Work}
While our work advances dialect-aware common sense reasoning, several limitations warrant discussion: the dialectal data generation process relied on GPT-4o for translation, which may introduce subtle semantic shifts or stylistic inconsistencies compared to naturally occurring dialectal speech, and while we implemented quality checks, the absence of large-scale human validation leaves room for potential noise, particularly in idiomatic expressions requiring deep cultural familiarity; the framework treats all dialects as equally distinct from MSA, overlooking gradient dialectal relationships. For instance, Levantine Arabic shares more lexical overlap with MSA than Moroccan Arabic, potentially leading to uneven generalization where linguistically closer dialects benefit implicitly; the binary labeling scheme (reasonable vs. non-reasonable) oversimplifies the continuum of common sense plausibility, failing to capture partially valid or context-dependent interpretations; moreover, the focus on four major dialects excludes dozens of other Arabic varieties, risking the marginalization of less common dialects like Sudanese or Yemeni Arabic, an area future work should address.


Future work will prioritize reducing the persistent Moroccan gap via dialect-specific adaptation (e.g., continued pretraining, normalization, or dialect-aware modules). We also plan to expand coverage to additional underrepresented varieties (e.g., Sudanese, Yemeni, Algerian, Iraqi), test broader diglossic/code-switched settings (e.g., Moroccan Arabic--French), and extend evaluation beyond sentence-level to contextual or multi-turn commonsense reasoning.

\section*{Ethical Statement}
\paragraph{Data License}{A primary ethical consideration in our work is the licensing and provenance of the data used. Our dataset builds upon two publicly available resources: the Arabic Dataset for Commonsense Validation and ArabicSense, both of which have been released for research purposes with appropriate usage permissions. To ensure compliance with licensing constraints, we generated novel dialectal variants derived from the Modern Standard Arabic (MSA) instances provided in the original datasets. This approach ensures that all newly created content remains consistent with the intended research scope of the original licenses and mitigates potential concerns related to data reuse and redistribution.}
\paragraph{Biased Language}{As the dialectal variants were generated using GPT-4o, we rely on the model's built-in safety mechanisms to not generate outputs that may contain biased, offensive, or contextually inappropriate language.}
\paragraph{Positive Impact of Commonsense Validation}{Our work advances existing methods and datasets for Arabic commonsense validation by introducing dialectal variants and exploring novel modeling approaches within this domain. We believe that enhancing commonsense understanding across Arabic dialects can contribute meaningfully to real-world applications such as fake news detection, fact-checking, and mitigating the spread of misleading or harmful content.}

\bibliographystyle{acl_natbib}
\bibliography{ref}

@article{tawalbeh2020sentence,
  author       = {Saja Khaled Tawalbeh and
                  Mohammad Al{-}Smadi},
  title        = {Is this sentence valid? An Arabic Dataset for Commonsense Validation},
  journal      = {CoRR},
  volume       = {abs/2008.10873},
  year         = {2020},
  url          = {https://arxiv.org/abs/2008.10873},
  eprinttype    = {arXiv},
  eprint       = {2008.10873},
  bibsource    = {dblp computer science bibliography, https://dblp.org}
}

@ARTICLE{scarselli-2008-GNNs,
  author={Scarselli, Franco and Gori, Marco and Tsoi, Ah Chung and Hagenbuchner, Markus and Monfardini, Gabriele},
  journal={IEEE Transactions on Neural Networks}, 
  title={The Graph Neural Network Model}, 
  year={2009},
  volume={20},
  number={1},
  pages={61-80},
  doi={10.1109/TNN.2008.2005605}}

@inproceedings{kipf-2016-GCNs,
  author       = {Thomas N. Kipf and
                  Max Welling},
  title        = {Semi-Supervised Classification with Graph Convolutional Networks},
  booktitle    = {5th International Conference on Learning Representations, {ICLR} 2017,
                  Toulon, France, April 24-26, 2017, Conference Track Proceedings},
  publisher    = {OpenReview.net},
  year         = {2017},
  url          = {https://openreview.net/forum?id=SJU4ayYgl},
  bibsource    = {dblp computer science bibliography, https://dblp.org}
}

@inproceedings{lamsiyah-etal-2025-arabicsense,
    title = "{A}rabic{S}ense: A Benchmark for Evaluating Commonsense Reasoning in {A}rabic with Large Language Models",
    author = "Lamsiyah, Salima  and
      Zeinalipour, Kamyar  and
      El amrany, Samir  and
      Brust, Matthias  and
      Maggini, Marco  and
      Bouvry, Pascal  and
      Schommer, Christoph",
    editor = "Ezzini, Saad  and
      Alami, Hamza  and
      Berrada, Ismail  and
      Benlahbib, Abdessamad  and
      El Mahdaouy, Abdelkader  and
      Lamsiyah, Salima  and
      Derrouz, Hatim  and
      Haddad Haddad, Amal  and
      Jarrar, Mustafa  and
      El-Haj, Mo  and
      Mitkov, Ruslan  and
      Rayson, Paul",
    booktitle = "Proceedings of the 4th Workshop on Arabic Corpus Linguistics (WACL-4)",
    month = jan,
    year = "2025",
    address = "Abu Dhabi, UAE",
    publisher = "Association for Computational Linguistics",
    url = "https://aclanthology.org/2025.wacl-1.1/",
    pages = "1--11"
}

@inproceedings{antoun-etal-2020-arabert,
    title = "{A}ra{BERT}: Transformer-based Model for {A}rabic Language Understanding",
    author = "Antoun, Wissam  and
      Baly, Fady  and
      Hajj, Hazem",
    editor = "Al-Khalifa, Hend  and
      Magdy, Walid  and
      Darwish, Kareem  and
      Elsayed, Tamer  and
      Mubarak, Hamdy",
    booktitle = "Proceedings of the 4th Workshop on Open-Source Arabic Corpora and Processing Tools, with a Shared Task on Offensive Language Detection",
    month = may,
    year = "2020",
    address = "Marseille, France",
    publisher = "European Language Resource Association",
    url = "https://aclanthology.org/2020.osact-1.2/",
    pages = "9--15",
    language = "eng",
    ISBN = "979-10-95546-51-1"
}

@inproceedings{abdul-mageed-etal-2021-arbert,
    title = "{ARBERT} {\&} {MARBERT}: Deep Bidirectional Transformers for {A}rabic",
    author = "Abdul-Mageed, Muhammad  and
      Elmadany, AbdelRahim  and
      Nagoudi, El Moatez Billah",
    editor = "Zong, Chengqing  and
      Xia, Fei  and
      Li, Wenjie  and
      Navigli, Roberto",
    booktitle = "Proceedings of the 59th Annual Meeting of the Association for Computational Linguistics and the 11th International Joint Conference on Natural Language Processing (Volume 1: Long Papers)",
    month = aug,
    year = "2021",
    address = "Online",
    publisher = "Association for Computational Linguistics",
    url = "https://aclanthology.org/2021.acl-long.551/",
    doi = "10.18653/v1/2021.acl-long.551",
    pages = "7088--7105"
}

@INPROCEEDINGS{10453697,
  author={Khaled, M Moneb and Sayadi, Aghyad Al and Elnagar, Ashraf},
  booktitle={2023 24th International Arab Conference on Information Technology (ACIT)}, 
  title={Commonsense Validation and Explanation in Arabic Text: A Comparative Study Using Arabic BERT Models}, 
  year={2023},
  volume={},
  number={},
  pages={1-6},
  doi={10.1109/ACIT58888.2023.10453697}}

@inproceedings{wang-etal-2020-semeval,
    title = "{S}em{E}val-2020 Task 4: Commonsense Validation and Explanation",
    author = "Wang, Cunxiang  and
      Liang, Shuailong  and
      Jin, Yili  and
      Wang, Yilong  and
      Zhu, Xiaodan  and
      Zhang, Yue",
    editor = "Herbelot, Aurelie  and
      Zhu, Xiaodan  and
      Palmer, Alexis  and
      Schneider, Nathan  and
      May, Jonathan  and
      Shutova, Ekaterina",
    booktitle = "Proceedings of the Fourteenth Workshop on Semantic Evaluation",
    month = dec,
    year = "2020",
    address = "Barcelona (online)",
    publisher = "International Committee for Computational Linguistics",
    url = "https://aclanthology.org/2020.semeval-1.39/",
    doi = "10.18653/v1/2020.semeval-1.39",
    pages = "307--321"
}

@inproceedings{lu-etal-2020-vgcnbert,
  title     = "{VGCN}-{BERT}: Augmenting {BERT} with Graph Embedding for Text Classification",
  author    = "Zhibin, Lu and Pan, Du and Jian-Yun, Nie",
  booktitle = "Proceedings of the 42nd European Conference on Information Retrieval (ECIR 2020)",
  year      = "2020",
  publisher = "Springer",
  organization = "European Conference on Information Retrieval",
  address   = "Lisbon, Portugal (Online)",
  url       = "http://rali.iro.umontreal.ca/rali/sites/default/files/publis/Lu2020_Chapter_VGCN-BERTAugmentingBERTWithGra.pdf",
  pages     = "369--382"
}

@inproceedings{Jiawei-etal-2020-graphbert,
  title     = "{GRAPH-BERT}: Only Attention is Needed for Learning Graph Representations",
  author    = "Jiawei, Zhang and Haopeng, Zhang and Congying, Xia and Li, Sun",
  journal   = "arXiv preprint arXiv:2001.05140",
  year      = "2020",
  url       = "https://arxiv.org/abs/2001.05140"
}

@inproceedings{sadallah-etal-2025-commonsense,
    title = "Commonsense Reasoning in {A}rab Culture",
    author = "Sadallah, Abdelrahman  and
      Tonga, Junior Cedric  and
      Almubarak, Khalid  and
      Almheiri, Saeed  and
      Atif, Farah  and
      Qwaider, Chatrine  and
      Kadaoui, Karima  and
      Shatnawi, Sara  and
      Alesh, Yaser  and
      Koto, Fajri",
    editor = "Che, Wanxiang  and
      Nabende, Joyce  and
      Shutova, Ekaterina  and
      Pilehvar, Mohammad Taher",
    booktitle = "Proceedings of the 63rd Annual Meeting of the Association for Computational Linguistics (Volume 1: Long Papers)",
    month = jul,
    year = "2025",
    address = "Vienna, Austria",
    publisher = "Association for Computational Linguistics",
    url = "https://aclanthology.org/2025.acl-long.380/",
    pages = "7695--7710",
    ISBN = "979-8-89176-251-0"
}

@article{openai2024gpt4o,
  author       = {OpenAI},
  title        = {GPT-4o System Card},
  journal      = {CoRR},
  volume       = {abs/2410.21276},
  year         = {2024},
  url          = {https://doi.org/10.48550/arXiv.2410.21276},
  doi          = {10.48550/ARXIV.2410.21276},
  eprinttype    = {arXiv},
  eprint       = {2410.21276},
  bibsource    = {dblp computer science bibliography, https://dblp.org}
}

@inproceedings{sap-etal-2020-commonsense,
    title = "Commonsense Reasoning for Natural Language Processing",
    author = "Sap, Maarten  and
      Shwartz, Vered  and
      Bosselut, Antoine  and
      Choi, Yejin  and
      Roth, Dan",
    editor = "Savary, Agata  and
      Zhang, Yue",
    booktitle = "Proceedings of the 58th Annual Meeting of the Association for Computational Linguistics: Tutorial Abstracts",
    month = jul,
    year = "2020",
    address = "Online",
    publisher = "Association for Computational Linguistics",
    url = "https://aclanthology.org/2020.acl-tutorials.7/",
    doi = "10.18653/v1/2020.acl-tutorials.7",
    pages = "27--33"
}

@inproceedings{ismayilzada-etal-2023-crow,
    title = "{CR}o{W}: Benchmarking Commonsense Reasoning in Real-World Tasks",
    author = "Ismayilzada, Mete  and
      Paul, Debjit  and
      Montariol, Syrielle  and
      Geva, Mor  and
      Bosselut, Antoine",
    editor = "Bouamor, Houda  and
      Pino, Juan  and
      Bali, Kalika",
    booktitle = "Proceedings of the 2023 Conference on Empirical Methods in Natural Language Processing",
    month = dec,
    year = "2023",
    address = "Singapore",
    publisher = "Association for Computational Linguistics",
    url = "https://aclanthology.org/2023.emnlp-main.607/",
    doi = "10.18653/v1/2023.emnlp-main.607",
    pages = "9785--9821"
}

@inproceedings{pires2019multilingual,
    title = "How Multilingual is Multilingual {BERT}?",
    author = "Pires, Telmo  and
      Schlinger, Eva  and
      Garrette, Dan",
    editor = "Korhonen, Anna  and
      Traum, David  and
      M{\`a}rquez, Llu{\'i}s",
    booktitle = "Proceedings of the 57th Annual Meeting of the Association for Computational Linguistics",
    month = jul,
    year = "2019",
    address = "Florence, Italy",
    publisher = "Association for Computational Linguistics",
    url = "https://aclanthology.org/P19-1493/",
    doi = "10.18653/v1/P19-1493",
    pages = "4996--5001"
}

@article{jiang2023mistral,
  author       = {Albert Q. Jiang and
                  Alexandre Sablayrolles and
                  Arthur Mensch and
                  Chris Bamford and
                  Devendra Singh Chaplot and
                  Diego de Las Casas and
                  Florian Bressand and
                  Gianna Lengyel and
                  Guillaume Lample and
                  Lucile Saulnier and
                  L{\'{e}}lio Renard Lavaud and
                  Marie{-}Anne Lachaux and
                  Pierre Stock and
                  Teven Le Scao and
                  Thibaut Lavril and
                  Thomas Wang and
                  Timoth{\'{e}}e Lacroix and
                  William El Sayed},
  title        = {Mistral 7B},
  journal      = {CoRR},
  volume       = {abs/2310.06825},
  year         = {2023},
  url          = {https://doi.org/10.48550/arXiv.2310.06825},
  doi          = {10.48550/ARXIV.2310.06825},
  eprinttype    = {arXiv},
  eprint       = {2310.06825},
  bibsource    = {dblp computer science bibliography, https://dblp.org}
}

@article{dubey2024llama,
  author       = {Hugo Touvron and
                  Thibaut Lavril and
                  Gautier Izacard and
                  Xavier Martinet and
                  Marie{-}Anne Lachaux and
                  Timoth{\'{e}}e Lacroix and
                  Baptiste Rozi{\`{e}}re and
                  Naman Goyal and
                  Eric Hambro and
                  Faisal Azhar and
                  Aur{\'{e}}lien Rodriguez and
                  Armand Joulin and
                  Edouard Grave and
                  Guillaume Lample},
  title        = {LLaMA: Open and Efficient Foundation Language Models},
  journal      = {CoRR},
  volume       = {abs/2302.13971},
  year         = {2023},
  url          = {https://doi.org/10.48550/arXiv.2302.13971},
  doi          = {10.48550/ARXIV.2302.13971},
  eprinttype    = {arXiv},
  eprint       = {2302.13971},
  bibsource    = {dblp computer science bibliography, https://dblp.org}
}

@inproceedings{Hwang2021COMETATOMIC2O,
  author       = {Jena D. Hwang and
                  Chandra Bhagavatula and
                  Ronan Le Bras and
                  Jeff Da and
                  Keisuke Sakaguchi and
                  Antoine Bosselut and
                  Yejin Choi},
  title        = {(Comet-) Atomic 2020: On Symbolic and Neural Commonsense Knowledge
                  Graphs},
  booktitle    = {Thirty-Fifth {AAAI} Conference on Artificial Intelligence, {AAAI}
                  2021, Thirty-Third Conference on Innovative Applications of Artificial
                  Intelligence, {IAAI} 2021, The Eleventh Symposium on Educational Advances
                  in Artificial Intelligence, {EAAI} 2021, Virtual Event, February 2-9,
                  2021},
  pages        = {6384--6392},
  publisher    = {{AAAI} Press},
  year         = {2021},
  url          = {https://doi.org/10.1609/aaai.v35i7.16792},
  doi          = {10.1609/AAAI.V35I7.16792},
  bibsource    = {dblp computer science bibliography, https://dblp.org}
}

@inproceedings{talmor-etal-2019-commonsenseqa,
    title = "{C}ommonsense{QA}: A Question Answering Challenge Targeting Commonsense Knowledge",
    author = "Talmor, Alon  and
      Herzig, Jonathan  and
      Lourie, Nicholas  and
      Berant, Jonathan",
    editor = "Burstein, Jill  and
      Doran, Christy  and
      Solorio, Thamar",
    booktitle = "Proceedings of the 2019 Conference of the North {A}merican Chapter of the Association for Computational Linguistics: Human Language Technologies, Volume 1 (Long and Short Papers)",
    month = jun,
    year = "2019",
    address = "Minneapolis, Minnesota",
    publisher = "Association for Computational Linguistics",
    url = "https://aclanthology.org/N19-1421/",
    doi = "10.18653/v1/N19-1421",
    pages = "4149--4158"
}

@inproceedings{10.1609/aaai.v33i01.33013027,
author = {Sap, Maarten and Le Bras, Ronan and Allaway, Emily and Bhagavatula, Chandra and Lourie, Nicholas and Rashkin, Hannah and Roof, Brendan and Smith, Noah A. and Choi, Yejin},
title = {ATOMIC: an atlas of machine commonsense for if-then reasoning},
year = {2019},
isbn = {978-1-57735-809-1},
publisher = {AAAI Press},
url = {https://doi.org/10.1609/aaai.v33i01.33013027},
doi = {10.1609/aaai.v33i01.33013027},
booktitle = {Proceedings of the Thirty-Third AAAI Conference on Artificial Intelligence and Thirty-First Innovative Applications of Artificial Intelligence Conference and Ninth AAAI Symposium on Educational Advances in Artificial Intelligence},
articleno = {372},
numpages = {9},
location = {Honolulu, Hawaii, USA},
series = {AAAI'19/IAAI'19/EAAI'19}
}

@inproceedings{10.5555/3031843.3031909,
  author       = {Hector J. Levesque and
                  Ernest Davis and
                  Leora Morgenstern},
  editor       = {Gerhard Brewka and
                  Thomas Eiter and
                  Sheila A. McIlraith},
  title        = {The Winograd Schema Challenge},
  booktitle    = {Principles of Knowledge Representation and Reasoning: Proceedings
                  of the Thirteenth International Conference, {KR} 2012, Rome, Italy,
                  June 10-14, 2012},
  publisher    = {{AAAI} Press},
  year         = {2012},
  url          = {http://www.aaai.org/ocs/index.php/KR/KR12/paper/view/4492},
  bibsource    = {dblp computer science bibliography, https://dblp.org}
}

@inproceedings{Bisk2020,
  author       = {Yonatan Bisk and
                  Rowan Zellers and
                  Ronan Le Bras and
                  Jianfeng Gao and
                  Yejin Choi},
  title        = {{PIQA:} Reasoning about Physical Commonsense in Natural Language},
  booktitle    = {The Thirty-Fourth {AAAI} Conference on Artificial Intelligence, {AAAI}
                  2020, The Thirty-Second Innovative Applications of Artificial Intelligence
                  Conference, {IAAI} 2020, The Tenth {AAAI} Symposium on Educational
                  Advances in Artificial Intelligence, {EAAI} 2020, New York, NY, USA,
                  February 7-12, 2020},
  pages        = {7432--7439},
  publisher    = {{AAAI} Press},
  year         = {2020},
  url          = {https://doi.org/10.1609/aaai.v34i05.6239},
  doi          = {10.1609/AAAI.V34I05.6239},
  bibsource    = {dblp computer science bibliography, https://dblp.org}
}

@inproceedings{sap-etal-2019-social,
    title = "Social {IQ}a: Commonsense Reasoning about Social Interactions",
    author = "Sap, Maarten  and
      Rashkin, Hannah  and
      Chen, Derek  and
      Le Bras, Ronan  and
      Choi, Yejin",
    editor = "Inui, Kentaro  and
      Jiang, Jing  and
      Ng, Vincent  and
      Wan, Xiaojun",
    booktitle = "Proceedings of the 2019 Conference on Empirical Methods in Natural Language Processing and the 9th International Joint Conference on Natural Language Processing (EMNLP-IJCNLP)",
    month = nov,
    year = "2019",
    address = "Hong Kong, China",
    publisher = "Association for Computational Linguistics",
    url = "https://aclanthology.org/D19-1454/",
    doi = "10.18653/v1/D19-1454",
    pages = "4463--4473"
}

@inproceedings{du-etal-2022-e,
    title = "e-{CARE}: a New Dataset for Exploring Explainable Causal Reasoning",
    author = "Du, Li  and
      Ding, Xiao  and
      Xiong, Kai  and
      Liu, Ting  and
      Qin, Bing",
    editor = "Muresan, Smaranda  and
      Nakov, Preslav  and
      Villavicencio, Aline",
    booktitle = "Proceedings of the 60th Annual Meeting of the Association for Computational Linguistics (Volume 1: Long Papers)",
    month = may,
    year = "2022",
    address = "Dublin, Ireland",
    publisher = "Association for Computational Linguistics",
    url = "https://aclanthology.org/2022.acl-long.33/",
    doi = "10.18653/v1/2022.acl-long.33",
    pages = "432--446"
}

@inproceedings{lin-etal-2020-commongen,
    title = "{C}ommon{G}en: A Constrained Text Generation Challenge for Generative Commonsense Reasoning",
    author = "Lin, Bill Yuchen  and
      Zhou, Wangchunshu  and
      Shen, Ming  and
      Zhou, Pei  and
      Bhagavatula, Chandra  and
      Choi, Yejin  and
      Ren, Xiang",
    editor = "Cohn, Trevor  and
      He, Yulan  and
      Liu, Yang",
    booktitle = "Findings of the Association for Computational Linguistics: EMNLP 2020",
    month = nov,
    year = "2020",
    address = "Online",
    publisher = "Association for Computational Linguistics",
    url = "https://aclanthology.org/2020.findings-emnlp.165/",
    doi = "10.18653/v1/2020.findings-emnlp.165",
    pages = "1823--1840"
}

@INPROCEEDINGS{9412167,
  author={Karimi, Akbar and Rossi, Leonardo and Prati, Andrea},
  booktitle={2020 25th International Conference on Pattern Recognition (ICPR)}, 
  title={Adversarial Training for Aspect-Based Sentiment Analysis with BERT}, 
  year={2021},
  volume={},
  number={},
  pages={8797-8803},
  doi={10.1109/ICPR48806.2021.9412167}}

@article{DBLP:journals/corr/abs-2108-13602,
  author       = {Javid Ebrahimi and
                  Hao Yang and
                  Wei Zhang},
  title        = {How Does Adversarial Fine-Tuning Benefit BERT?},
  journal      = {CoRR},
  volume       = {abs/2108.13602},
  year         = {2021},
  url          = {https://arxiv.org/abs/2108.13602},
  eprinttype    = {arXiv},
  eprint       = {2108.13602},
  bibsource    = {dblp computer science bibliography, https://dblp.org}
}

@inproceedings{alshahrani-etal-2024-arabic,
    title = "{A}rabic Synonym {BERT}-based Adversarial Examples for Text Classification",
    author = "Alshahrani, Norah  and
      Alshahrani, Saied  and
      Wali, Esma  and
      Matthews, Jeanna",
    editor = "Falk, Neele  and
      Papi, Sara  and
      Zhang, Mike",
    booktitle = "Proceedings of the 18th Conference of the European Chapter of the Association for Computational Linguistics: Student Research Workshop",
    month = mar,
    year = "2024",
    address = "St. Julian{'}s, Malta",
    publisher = "Association for Computational Linguistics",
    url = "https://aclanthology.org/2024.eacl-srw.10/",
    doi = "10.18653/v1/2024.eacl-srw.10",
    pages = "137--147"
}

@inproceedings{Loshchilov2017DecoupledWD,
  title={Decoupled Weight Decay Regularization},
  author={Ilya Loshchilov and Frank Hutter},
  booktitle={International Conference on Learning Representations},
  year={2017},
  url={https://api.semanticscholar.org/CorpusID:53592270}
}

@article{10.1145/2701413,
author = {Davis, Ernest and Marcus, Gary},
title = {Commonsense reasoning and commonsense knowledge in artificial intelligence},
year = {2015},
issue_date = {September 2015},
publisher = {Association for Computing Machinery},
address = {New York, NY, USA},
volume = {58},
number = {9},
issn = {0001-0782},
url = {https://doi.org/10.1145/2701413},
doi = {10.1145/2701413},
journal = {Commun. ACM},
month = aug,
pages = {92–103},
numpages = {12}
}

@inproceedings{lin-etal-2019-kagnet,
    title = "{K}ag{N}et: Knowledge-Aware Graph Networks for Commonsense Reasoning",
    author = "Lin, Bill Yuchen  and
      Chen, Xinyue  and
      Chen, Jamin  and
      Ren, Xiang",
    editor = "Inui, Kentaro  and
      Jiang, Jing  and
      Ng, Vincent  and
      Wan, Xiaojun",
    booktitle = "Proceedings of the 2019 Conference on Empirical Methods in Natural Language Processing and the 9th International Joint Conference on Natural Language Processing (EMNLP-IJCNLP)",
    month = nov,
    year = "2019",
    address = "Hong Kong, China",
    publisher = "Association for Computational Linguistics",
    url = "https://aclanthology.org/D19-1282/",
    doi = "10.18653/v1/D19-1282",
    pages = "2829--2839"
}

@article{hochreiter1997long,
author = {Hochreiter, Sepp and Schmidhuber, J\"{u}rgen},
title = {Long Short-Term Memory},
year = {1997},
issue_date = {November 15, 1997},
publisher = {MIT Press},
address = {Cambridge, MA, USA},
volume = {9},
number = {8},
issn = {0899-7667},
url = {https://doi.org/10.1162/neco.1997.9.8.1735},
doi = {10.1162/neco.1997.9.8.1735},
journal = {Neural Comput.},
month = nov,
pages = {1735–1780},
numpages = {46}
}

@misc{ganin2016domainadversarialtrainingneuralnetworks,
      title={Domain-Adversarial Training of Neural Networks}, 
      author={Yaroslav Ganin and Evgeniya Ustinova and Hana Ajakan and Pascal Germain and Hugo Larochelle and François Laviolette and Mario Marchand and Victor Lempitsky},
      year={2016},
      eprint={1505.07818},
      archivePrefix={arXiv},
      primaryClass={stat.ML},
      url={https://arxiv.org/abs/1505.07818}, 
}

@inproceedings{bhatia-etal-2025-swan,
    title = "Swan and {A}rabic{MTEB}: Dialect-Aware, {A}rabic-Centric, Cross-Lingual, and Cross-Cultural Embedding Models and Benchmarks",
    author = "Bhatia, Gagan  and
      Nagoudi, El Moatez Billah  and
      El Mekki, Abdellah  and
      Alwajih, Fakhraddin  and
      Abdul-Mageed, Muhammad",
    editor = "Chiruzzo, Luis  and
      Ritter, Alan  and
      Wang, Lu",
    booktitle = "Findings of the Association for Computational Linguistics: NAACL 2025",
    month = apr,
    year = "2025",
    address = "Albuquerque, New Mexico",
    publisher = "Association for Computational Linguistics",
    url = "https://aclanthology.org/2025.findings-naacl.263/",
    doi = "10.18653/v1/2025.findings-naacl.263",
    pages = "4654--4670",
    ISBN = "979-8-89176-195-7"
}

@inproceedings{shaban-habash-2025-arabic,
    title = "The {A}rabic Generality Score: Another Dimension of Modeling {A}rabic Dialectness",
    author = "Sha{'}ban, Sanad  and
      Habash, Nizar",
    editor = "Christodoulopoulos, Christos  and
      Chakraborty, Tanmoy  and
      Rose, Carolyn  and
      Peng, Violet",
    booktitle = "Proceedings of the 2025 Conference on Empirical Methods in Natural Language Processing",
    month = nov,
    year = "2025",
    address = "Suzhou, China",
    publisher = "Association for Computational Linguistics",
    url = "https://aclanthology.org/2025.emnlp-main.1524/",
    doi = "10.18653/v1/2025.emnlp-main.1524",
    pages = "29990--30001",
    ISBN = "979-8-89176-332-6"
}

@inproceedings{inoue-etal-2021-interplay,
    title = "The Interplay of Variant, Size, and Task Type in {A}rabic Pre-trained Language Models",
    author = "Inoue, Go  and
      Alhafni, Bashar  and
      Baimukan, Nurpeiis  and
      Bouamor, Houda  and
      Habash, Nizar",
    editor = "Habash, Nizar  and
      Bouamor, Houda  and
      Hajj, Hazem  and
      Magdy, Walid  and
      Zaghouani, Wajdi  and
      Bougares, Fethi  and
      Tomeh, Nadi  and
      Abu Farha, Ibrahim  and
      Touileb, Samia",
    booktitle = "Proceedings of the Sixth Arabic Natural Language Processing Workshop",
    month = apr,
    year = "2021",
    address = "Kyiv, Ukraine (Virtual)",
    publisher = "Association for Computational Linguistics",
    url = "https://aclanthology.org/2021.wanlp-1.10/",
    pages = "92--104"
}

\clearpage
\appendix

\renewcommand{\thesection}{\Alph{section}}
\renewcommand{\thesubsection}{\thesection.\arabic{subsection}}

\section*{Appendix}

\section{Related Work}
\label{sec:relatedwork}

\subsection{Commonsense Reasoning Datasets}

\paragraph{Common Sense Reasoning in English} 
There have been many benchmarks for English commonsense reasoning, such as 
CommonSenseQA ~\citep{talmor-etal-2019-commonsenseqa}, ComVe~\citep{wang-etal-2020-semeval}, ATOMIC~\citep{10.1609/aaai.v33i01.33013027}  and ATOMIC 2020~\citep{Hwang2021COMETATOMIC2O}. Within the broader scope of commonsense reasoning, several specialized subfields have emerged, each targeting distinct types of implicit human knowledge.
Earlier work focused on pronoun coreference resolution in linguistic contexts~\citep{10.5555/3031843.3031909}, physical commonsense reasoning~\citep{Bisk2020}, social reasoning~\citep{sap-etal-2019-social}, and causal reasoning~\citep{du-etal-2022-e}. Additional efforts have explored commonsense in natural language generation~\citep{lin-etal-2020-commongen}, as well as the integration of commonsense reasoning into real-world NLP tasks~\citep{ismayilzada-etal-2023-crow}.

Despite these advancements, most research and benchmarks are centered around English, leaving many other languages, such as Arabic, under-resourced.

\paragraph{CommonSense Reasoning in Arabic} 
Recent years have witnessed exploration of Arabic commonsense reasoning. Initial efforts focused on 
translating English commonsense benchmarks into Modern Standard Arabic (MSA) \citep{tawalbeh2020sentence}, or leveraging large language models (LLMs) to generate MSA data from seed data \citep{lamsiyah-etal-2025-arabicsense}. However, these datasets lack cultural nuances of Arabic. Recent work by \citet{sadallah-etal-2025-commonsense} fills this gap by collecting a dataset 
covering cultures of 13 countries across the Gulf, Levant, North Africa, and the Nile Valley. Despite this advancement, their dataset remains restricted to MSA and does not encompass the rich linguistic and cultural diversity embedded in Arabic dialects.
Therefore, we collect the first Arabic dialects commonsense reasoning benchmark, extending commonsense evaluation to Arabic dialects, aiming to capture more authentic and regionally grounded reasoning patterns.

\subsection{Approaches for Commonsense Reasoning}
Prior research has primarily focused on fine-tuning transformer-based models or employing LLMs for commonsense validation and explanation generation, without introducing improved task-specific representations that could enhance performance. \citet{tawalbeh2020sentence} fine-tuned BERT, USE, and ULMFit models for binary classification, selecting the more plausible sentence from a pair. More recently, \citet{lamsiyah-etal-2025-arabicsense} evaluated a suite of BERT-based encoders, including AraBERTv2 \citep{antoun-etal-2020-arabert}, ARBERT, MARBERTv2 \citep{abdul-mageed-etal-2021-arbert}, CaMeLBERT, and mBERT \citep{pires2019multilingual}, on two classification tasks: \emph{(i)} distinguishing commonsensical from nonsensical statements, and \emph{(ii)} identifying the underlying reasoning behind nonsensicality. They also assessed causal LLMs including Mistral-7B \citep{jiang2023mistral}, LLaMA-3 \citep{dubey2024llama} and Gemma, on the two tasks above, along with the task \emph{(iii)} generating natural language explanations for commonsense violations. 
These approaches lacked exploring better representation learning techniques to enhance the performance. 

\paragraph{Integration of Adversarial Training with Encoder Transformer Models} 
Prior work has explored integrating adversarial training with transformer-based models.
\citet{9412167} introduced BERT Adversarial Training (BAT), which fine-tuned BERT and domain-specific BERT-PT using adversarial perturbations in the embedding space to improve robustness in Aspect-Based Sentiment Analysis (ABSA). 
\citet{DBLP:journals/corr/abs-2108-13602} showed that adversarial training can preserve BERT's syntactic abilities, such as word order sensitivity and parsing, during fine-tuning, compared to standard fine-tuning. Additionally, it demonstrated how adversarial training prevented BERT from oversimplifying representations by reducing over-reliance on a few words, leading to better generalization. 

In Arabic context, \citet{alshahrani-etal-2024-arabic} conducted a synonym-based word-level adversarial attack on Arabic text classification models using a Masked Language Modeling (MLM) task with AraBERT. This attack replaces important words in the input text with semantically similar synonyms predicted by AraBERT to generate adversarial examples that can fool state-of-the-art classifiers. To ensure grammatical correctness, they utilize CAMeLBERT as a Part-of-Speech tagger to verify that the synonym replacements match the original word’s grammatical tags, maintaining sentence grammar.
%
 
We investigate the use of adversarial training across dialects as a means to learn more robust and generalized representations, thereby enhancing model performance and resilience across the diverse landscape of Arabic dialects.

\paragraph{Integration of Graph-based Approaches with Encoder Transformer Models}
Graph Neural Networks (GNNs)~\citep{scarselli-2008-GNNs}, and particularly Graph Convolutional Networks (GCNs)~\citep{kipf-2016-GCNs} have gained significant attention for their ability to model relational and topological structures in data. 
Integrating graph-based structures with encoder-based Transformer models enables models to better grasp higher-level connections and contextual dependencies that are crucial for complex language understanding tasks like commonsense reasoning. 
For example, 
GraphBERT~\citep{Jiawei-etal-2020-graphbert} introduced leveraging Transformer-style self-attention over linkless subgraphs, allowing it to learn graph representations without relying on explicit edge connections. This approach mitigates issues such as over-smoothing and enhances parallelizability. In contrast, VGCN-BERT~\citep{lu-etal-2020-vgcnbert} adopts a hybrid design, incorporating a vocabulary-level graph convolutional network (VGCN) into the BERT architecture. It constructs a global word co-occurrence graph and fuses the GCN-derived word representations with the BERT input embeddings, thereby enriching the model’s understanding of global corpus-level semantics. Both models demonstrate how graph-derived features, when fused effectively with Transformer encoders, can improve downstream tasks like text classification by fusing graph extracted morphological features with the token-level contextual embeddings. In the context of commonsense reasoning, \citep{lin-etal-2019-kagnet} proposed KAGNet, a model that integrates GCNs with Long Short-Term Memory networks (LSTMs)~\citep{hochreiter1997long} to encode knowledge paths from external commonsense knowledge bases, thereby improving question answering performance through structured reasoning.

In this work, we integrate graph neural network-projected embeddings into transformer-based encoders, enriching contextual representations with global structural information which are critical to common sense validation. 



\section{Data Generation and Quality Control}
\label{datagen}
\paragraph{Dialect generation} We design a prompt tailored for accurate and meaning-preserved translation: 
\begin{quote}
    \small
    \begin{RLtext}
            \texttt{أنت خبير في اللهجات العربية. ترجم الجملة التالية إلى اللهجة \LR{\{dialect\}} بدون تغيير المعنى:\LR{\{sentence\}}}.
    \end{RLtext}
\end{quote}  
The prompt translates to “You are an expert in Arabic dialects. Translate the following sentence to \{dialect\}: \{sentence\}”. 
This ensures that the intended meaning of each sentence remains intact while reflecting natural dialectal usage.

\paragraph{Automatic Evaluation} We prompt Gemini 2.5 Flash to assess the faithfulness of dialectal translations as follows:

\begin{quote}
\small
\textbf{System Prompt:} You are an expert Arabic linguist. Your task is to verify whether \texttt{DIALECT\_TEXT} is a correct translation of \texttt{MSA\_TEXT} from Modern Standard Arabic (MSA) into the specified Arabic dialect. A translation is correct if the meaning, events, entities, time references, and polarity in \texttt{MSA\_TEXT} are faithfully preserved in \texttt{DIALECT\_TEXT}, even if wording differs due to dialectal variation. Ignore minor spelling, punctuation, and orthographic differences. Do not allow additions, omissions, or factual changes. Output only one word: \texttt{true} or \texttt{false}. Do not explain your decision.

\vspace{0.5em}

\texttt{DIALECT: <dialect\_name>}\\
\texttt{MSA\_TEXT: <msa\_text>}\\
\texttt{DIALECT\_TEXT: <dialect\_text>}\\
\texttt{Answer:}
\end{quote}

\section{Graph Embeddings-based Encoder Transformer models}
\label{append:graph}
\subsection{Extended Explanation of methodology}
\begin{figure*}[t]
  \centering
  \includegraphics[width=\textwidth]{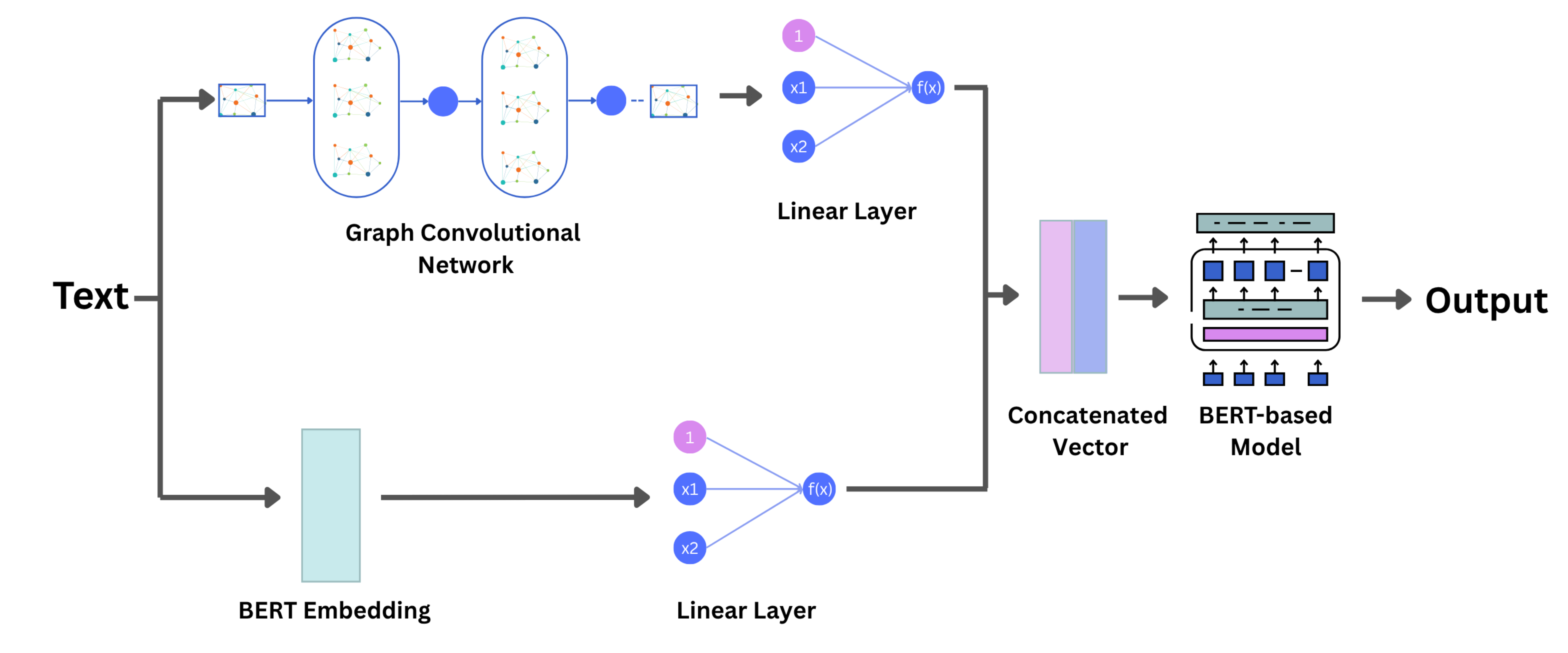}
  \caption{BERT Model with Graph Embeddings Fusion.}
  \label{fig:model_arch}
\end{figure*}

Figure \ref{fig:model_arch} and Algorithm \ref{algo:training-graphTransformer} summarize the pipeline.

\begin{algorithm}[h!]
\caption{The training algorithm for Graph Embeddings-based Encoder Transformer models.}
\label{algo:training-graphTransformer}
\begin{algorithmic}[1]

\State \textbf{Given:}
\State $\mathcal{D}_{\text{train}}$ \Comment{Labeled corpora of text samples}
\State $\mathcal{T}$ \Comment{Pretrained textual encoder (e.g., BERT)}
\State $\mathcal{G}$ \Comment{Graph encoder (e.g., GCN)}
\State $\mathcal{F}$ \Comment{Fusion mechanism (e.g., attention)}
\State $\mathcal{C}$ \Comment{Classifier head}
\State $\theta$ \Comment{Trainable parameters}

\vspace{1mm}
\State \textbf{Preparation:}
\ForAll{$(x, y) \in \mathcal{D}$}
    \State Tokenize $x \rightarrow$ $\mathbf{t} \in \mathbb{R}^{L \times h}$
    \State Convert $x \rightarrow$ graph $\mathcal{G}_x = (V, E, \mathbf{X})$
\EndFor

\vspace{1mm}
\State \textbf{Initialize:}
\State $\theta \gets$ random or pretrained weights

\vspace{1mm}
\For{$e = 1$ to $E$}
    \State \textbf{Training Step:}
    \ForAll{$(x, y, \mathcal{G}_x) \in \mathcal{D}_{\text{train}}$}
        \State $\mathbf{z}_t \gets \mathcal{T}(x)$ \Comment{Textual representation}
        \State $\mathbf{z}_g \gets \mathcal{G}(\mathcal{G}_x)$ \Comment{Graph representation}
        \State $\mathbf{z}_f \gets \mathcal{F}(\mathbf{z}_t, \mathbf{z}_g)$ \Comment{Fusion}
        \State $\hat{y} \gets \mathcal{C}(\mathbf{z}_f)$ \Comment{Prediction}
        \State Update $\theta$ via $\nabla_{\theta} \mathcal{L}(\hat{y}, y)$
    \EndFor

    \vspace{1mm}

\EndFor

\vspace{1mm}

\end{algorithmic}
\end{algorithm}

\section{Domain Adversarial Training}
\label{append:adversarial}

\subsection{Method}
\label{sec:adversarial-meth}
We implement domain-adversarial training (DANN)~\citep{ganin2016domainadversarialtrainingneuralnetworks} to encourage dialect-invariant features during fine-tuning. A shared Transformer encoder produces a sentence representation $\mathbf{h}$ from the final-layer \texttt{[CLS]} vector. We attach two MLP heads: (i) a task classifier for commonsense validation, and (ii) a dialect discriminator predicting the dialect label. The dialect discriminator receives $\mathbf{h}$ through a Gradient Reversal Layer (GRL), which multiplies its backpropagated gradient by $-\alpha$, so the encoder is optimized to reduce task loss while making dialect prediction harder.

Both heads use the same architecture:
\texttt{Dropout} $\rightarrow$ \texttt{Linear}($H\!\rightarrow\!768$) $\rightarrow$ \texttt{ReLU} $\rightarrow$
\texttt{Dropout} $\rightarrow$ \texttt{Linear}($768\!\rightarrow\!C$),
with dropout rate $0.1$, where $C{=}2$ for the main task and $C{=}5$ for dialect prediction.

We optimize the combined objective:
\begin{equation}
\mathcal{L}=\mathcal{L}_{\text{main}}+\lambda\,\mathcal{L}_{\text{dial}},
\end{equation}
where both terms are cross-entropy losses and the GRL applies the adversarial signal to the shared encoder. We set $\lambda{=}1.0$ and use a simple schedule for the GRL strength $\alpha=\min(1,\frac{e+1}{5})$ as a function of epoch index $e$.

\subsection{Experimental Setup}
We follow the same data split and base fine-tuning setup described in Section~\ref{sec:exp-setup}. Adversarial models are trained for 3 epochs using AdamW (learning rate $2e-5$, weight decay $0.01$), with gradient clipping (max norm $1.0$) and a linear warmup over the first 100 optimization steps (from $0.1\times$lr to lr, then constant). We select the best checkpoint using the dev weighted F1 of the main task and report results on the held-out test set.

\subsection{Results}

Table \ref{tab:adv-acc-results} compares the baseline models against their domain-adversarial training variants. Overall, domain-adversarial training consistently leads to a substantial degradation in accuracy, indicating that enforcing domain invariance in this setting harms learning rather than improving cross-dialect generalization.

\begin{table*}[t!]
\centering
\small
\begin{tabular}{lcccccc}
\toprule
\textbf{Methods} & \textbf{MSA$\uparrow$} & \textbf{Egyptian$\uparrow$} & \textbf{Gulf$\uparrow$} & \textbf{Levantine$\uparrow$} & \textbf{Moroccan$\uparrow$} & \textbf{Avg.$\uparrow$} \\ 
\midrule
AraBERTv2  & \textbf{65.88} & \textbf{68.33}& \textbf{65.78}& \textbf{65.36}& \textbf{62.09}& \textbf{65.34} \\
AraBERTv2 (Adv) & 50.52& 50.15& 49.89& 50.03& 50.55& 50.23 \\
\midrule
CAMeLBERT-mix &  \textbf{73.30} & \textbf{73.52} & \textbf{74.63}& \textbf{73.82} & \textbf{66.45} & \textbf{72.34} \\
CAMeLBERT-mix (Adv) & 66.52& 67.12& 67.26& 68.91& 64.18 & 66.80 \\
\midrule
MARBERTv2  & \textbf{80.12}& \textbf{80.73}& \textbf{78.81}& \textbf{80.03}& \textbf{71.09} & \textbf{78.16} \\
MARBERTv2 (Adv) & 79.58& 79.00& 77.57& 78.52& 70.24& 76.98 \\

\bottomrule
\end{tabular}
\caption{Accuracy (\%) of baselines and adversarial training-based models on MSA and dialects datasets.} 
\label{tab:adv-acc-results}
\end{table*}

\paragraph{Why Does Adversarial Training Fail?}
All three backbones degrade under adversarial training, and the degradation is especially severe for AraBERTv2 (dropping to near chance). This pattern strongly suggests that dialect-specific cues are \emph{not} purely nuisance variation for this task: enforcing dialect invariance likely removes information that is genuinely predictive (lexical, morphological, or orthographic markers correlated with the label). Another plausible contributor is optimization instability: if the adversarial signal is too strong relative to the supervised signal, feature collapse can occur. An effect that would be amplified for a less dialect-robust backbone such as AraBERTv2. Practically, these results argue against treating dialect simply as a domain to be ``erased''; a better direction may be \emph{domain-aware} modeling (e.g., dialect embeddings/adapters or mixture-of-experts) rather than domain-invariant representations.
In contrast to adversarial training, adding GCN-based embeddings improves \emph{every} backbone. 


\end{document}